%% file: main.tex
\documentclass[letterpaper,10pt,conference]{ieeeconf}

\IEEEoverridecommandlockouts 

\usepackage[T1]{fontenc}
\usepackage[utf8]{inputenc}
\usepackage{mathptmx}
\usepackage{amsmath,amssymb,bm}
\usepackage{booktabs,array}
\usepackage{placeins} 
\usepackage{cite,graphicx,microtype}

\newsavebox{\marginfigbox}
\newcommand{\marginSafeFigure}[2][\columnwidth]{%
  \sbox{\marginfigbox}{\includegraphics[width=#1]{#2}}%
  \raisebox{-9pt}[\ht\marginfigbox][\dp\marginfigbox]{%
    \usebox{\marginfigbox}}%
}

\usepackage[hidelinks]{hyperref}
\title{Family-Air: Wind-Aware Trajectory-Family Optimization for Urban Aerial Transport in Virtual Tubes}
\author{Haozhe Wang$^{1}$ and Xun Huang$^{1}$%
\thanks{$^{1}$ Haozhe Wang and Xun Huang are with the
School of Mechanics and Engineering Science, Peking University, Beijing, 100871, P.R. China
(e-mail: \texttt{2501111813@stu.pku.edu.cn; huangxun@pku.edu.cn).}}%
}

\begin{document}
\maketitle
\begin{abstract}\input{abstract}\end{abstract}
\input{sections/introduction}
\input{sections/related_work}

\input{sections/formulation}
\input{figures/workflow}
\input{sections/method}

\input{sections/experiments}
\input{sections/conclusion}

\appendix

The safe-sphere roadmap is searched
using A* with an augmented state
representation to account for changes
in heading. Each state $(h,m)$ consists
of the predecessor sphere $h$ and
the current sphere $m$. A transition to a neighboring sphere $n$ updates the state to $(m,n)$ and incurs the incremental cost
\begin{equation}
\begin{aligned}
\Delta J(h,m,n)
={}&d_{mn}\left[
\lambda_L+\lambda_B
\left(\frac{r_0}{b_{mn}}\right)^2
\right]\\
&+\lambda_C+
\lambda_\theta\left(
1-\hat d_{hm}^{\top}\hat d_{mn}
\right).
\end{aligned}
\label{eq:routecost}
\end{equation}

The sphere centers are $c_i$, with $d_{ij}=\|c_j-c_i\|$ and $\hat d_{ij}=(c_j-c_i)/d_{ij}$. Let $R_i=r_i-r_{\mathrm{form}}$ be the effective
core radius, where
$r_{\mathrm{form}}=\max_{1\leq j\leq N}\|E(0)\xi_j\|$
is the radius of the initial formation and
$r_i$ already incorporates the safety margins. For adjacent spheres $m$ and $n$, the bottleneck radius is computed as
\begin{equation}
\begin{aligned}
s_{mn}&=\operatorname{clip}\left(
\frac{d_{mn}^2+R_m^2-R_n^2}{2d_{mn}},
0,d_{mn}\right),\\
b_{mn}&=\sqrt{\max\left\{
R_m^2-s_{mn}^2,\,
R_n^2-(d_{mn}-s_{mn})^2,\,
0
\right\}}.
\end{aligned}
\label{eq:bottleneck}
\end{equation}
This radius characterizes the narrowest cross-section of the union of the two effective cores along the line segment connecting their centers. The scaling parameter $r_0$ normalizes the bottleneck penalty. The nonnegative weights $\lambda_L$, $\lambda_B$, $\lambda_C$, and $\lambda_\theta$ penalize travel distance, narrow connections, inter-sphere transitions, and turning changes.

\bibliographystyle{IEEEtran}
\bibliography{references}
\end{document}

%% file: abstract.tex
Coordinated aerial transport in dense urban environments requires drone formations to navigate confined spaces and adapt to spatially varying winds around buildings. Formation deformation enables passage through narrow spaces but changes individual drone velocities and the local winds they encounter, coupling formation motion with aerodynamic effects. We propose Family-Air, a wind-aware trajectory-family planner that constructs a virtual tube as shared collision-free flight space and jointly optimizes formation-center motion, deformation, and timing. A shared affine transformation represents a continuous trajectory family, incorporating spatially varying wind and deformation-induced velocities into a unified air-relative motion objective. The optimization enforces virtual-tube containment, inter-drone separation, and kinematic feasibility. Simulations across four urban wind-field scenarios demonstrate the importance of local wind–deformation coupling. Flight simulations further show improvements in trajectory tracking accuracy. These results demonstrate the potential of wind-aware trajectory-family planning for coordinated aerial transport in complex urban environments.

%% file: sections/introduction.tex
\section{Introduction}\label{sec:intro}

Urban aerial transport is attracting growing
attention, with the safety of low-altitude
operations becoming a key challenge for wider
deployment~\cite{huang2025small,pang2022risk}.
Coordinated aerial delivery can involve
multiple drones flying in formation along
shared routes~\cite{alkouz2020formation}. Such missions require obstacle avoidance, coordinated motion, and safe separation among drones~\cite{quan2023formation}. However, a path that is collision-free for a single drone may not accommodate an entire formation. Navigating narrow gaps or turns between buildings may require the formation to contract or rotate. 
Meanwhile, urban buildings create spatially varying wind fields, exposing drones within the same formation to different local wind conditions. These winds alter the air-relative velocities of individual drones, affecting aerodynamic loading and potentially influencing propulsion power~\cite{zeng2019energy,bauersfeld2022range}.
Such wind conditions can also complicate trajectory tracking and coordinated flight~\cite{folk2025planning}.
Consequently, urban formation trajectory planning must address both the geometric constraints of confined flight space and the effects of spatially varying winds.
\input{figures/method}

To address the geometric constraints of flight in cluttered environments, safe-corridor methods incorporate obstacle avoidance into continuous trajectory optimization and provide an effective approach to motion planning~\cite{liu2017corridor}. For multi-drone systems, virtual-tube methods further provide shared flight space for coordinated formation motion and support the construction of trajectory families~\cite{mao2024tube}. These methods establish a geometric basis for coordinated passage, but identifying feasible flight space does not determine how the formation should move within it. In particular, under spatially varying winds, the formation must still adapt its trajectory, shape, and flight speed to local conditions.

Existing studies have addressed coordinated formation motion and deformation in cluttered environments~\cite{quan2023formation,sha2024affine}, but do not explicitly account for spatially varying winds during trajectory optimization. Wind-aware planning studies have incorporated spatial wind information into path selection~\cite{pensado2024turbulence,veiga2025hybrid}, local motion planning~\cite{folk2025planning}, and trajectory and speed optimization~\cite{qin2026geowind2plan}. 
However, these wind-aware methods primarily focus on individual drones, with limited consideration of deformable formations.

In spatially varying wind, formation deformation
affects each drone's air-relative velocity
by changing both its ground-relative velocity
and the position at which it samples the
wind field, as illustrated in
Fig.~\ref{fig:method}.
Evaluating wind only at the formation center
neglects spatial wind variations across
the formation and may therefore misestimate
the air-relative motion cost.

We propose Family-Air (FA), a wind-aware trajectory-family planner to address this problem. The planner first automatically constructs a virtual tube from the start, goal, and surrounding obstacles, then jointly optimizes formation-center motion, deformation, and flight time within the tube. A time-varying formation center and a shared affine deformation map represent a continuous family of trajectories. The air-relative motion cost is evaluated over the entire family, accounting for deformation-induced drone velocities and position-dependent local wind. The optimization balances air-relative motion cost,
trajectory smoothness, flight time, and deformation
regularization, subject to virtual-tube containment,
inter-drone separation, ground-relative kinematic bounds,
and sampled air-relative speed limits.

The main contributions of this work are as follows:
\begin{enumerate}
\setlength{\itemsep}{2pt}
\setlength{\parsep}{0pt}
\item \textbf{Wind--deformation coupling formulation.} We formulate an air-relative motion cost over a continuous trajectory family that jointly accounts for deformation-induced changes in drone velocities and position-dependent local wind.
\item \textbf{Tube-constrained joint trajectory optimization.} We jointly optimize formation-center motion, deformation, and timing within an automatically constructed virtual tube, enforcing continuous geometric and kinematic constraints and  sampled air-relative
speed limits.
\item \textbf{Urban wind-field simulation and validation.} We evaluate the method through benchmark and ablation studies in three urban wind-field scenes and a Chicago-derived heightfield scene, examining the performance of wind-aware planning.
\end{enumerate}

%% file: figures/method.tex
\begin{figure}[t]
\centering
\includegraphics[width=\columnwidth]{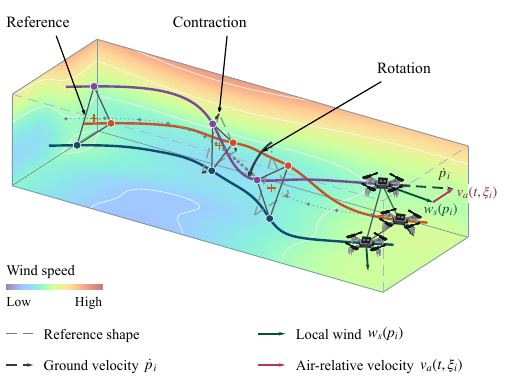}
\caption{Wind--deformation coupling in a spatially varying wind field.
Formation contraction and rotation change
both drone velocities and the positions at
which local wind is sampled.}
\label{fig:method}
\end{figure}

%% file: sections/related_work.tex
\section{Related Work}\label{sec:related}
\subsection{Safe Corridors and Virtual Tubes}

Safe-corridor methods constrain trajectories to convex collision-free regions, simplifying obstacle avoidance in trajectory optimization. For example, Liu et al. combine safe-flight corridors with piecewise-polynomial optimization to generate dynamically feasible trajectories in 3D environments~\cite{liu2017corridor}. Ren et al. use overlapping spherical corridors with receding updates for high-speed flight in unknown environments~\cite{ren2022bubble}. Park et al. extend linear safe corridors to distributed multi-drone trajectory optimization with inter-drone collision avoidance and dynamic feasibility~\cite{park2022lsc}.

By contrast, virtual-tube methods offer another planning and control framework for organizing group motion within shared flight space. Mao and Quan develop regular virtual tubes by constraining tube surfaces, cross sections, and geometric continuity~\cite{mao2022regular}. Quan et al. study distributed control for robots passing through straight and curved tubes, addressing tube containment and collision avoidance~\cite{quan2023curve,quan2022vtol}. For trajectory planning, Mao et al. show
that a continuous trajectory family can
be constructed from convex combinations
of finitely many optimal trajectories
under specified conditions~\cite{mao2024tube}. Tube RRT* considers passage volume and path length in homotopic path search to select group routes with greater spatial clearance~\cite{mao2025tuberrt}. In this work, we further combine virtual-tube containment with wind-aware optimization of a shared affine trajectory family.

\subsection{Formation and Deformation Planning}

Low-dimensional representations describe coordinated formation motion through shared variables. Zhao develops an affine formation maneuver control framework that enables formation translation, rotation, scaling, and deformation~\cite{zhao2018affine}. In cluttered environments, Alonso-Mora et al. optimize formation parameters under collision constraints~\cite{alonsomora2017formation}, and Quan et al. combine formation similarity, adaptive reorganization, obstacle avoidance, and dynamic feasibility in distributed trajectory optimization~\cite{quan2023formation}. Sha et al. further perform global trajectory search and optimization directly in the space of affine formation parameters~\cite{sha2024affine}. Our method represents formation motion and deformation through shared affine variables, allowing the trajectories of all drones to be optimized within a compact parameterization.

\subsection{Wind-Aware Path and Trajectory Planning}

Wind-field estimation provides spatial wind information for low-altitude planning. 
Folk et al. estimate local urban airflow using building geometry and sparse observations~\cite{folk2024learning}. At the path-planning level, several studies account for turbulence and wind risk in route selection~\cite{pensado2024turbulence,veiga2025hybrid}. Gu et al. combine computational fluid dynamics (CFD) wind fields with wind-aware path search and flight-speed adjustment for urban logistics drones~\cite{gu2026windfield}. Beyond route selection, wind information has also been incorporated into formation configuration selection and continuous trajectory optimization. Alkouz and Bouguettaya investigate fixed and adaptive formation configurations under varying wind conditions along delivery routes~\cite{alkouz2020formation}. For single-drone trajectory planning, Habib et al.
integrate rapid CFD wind-field generation
with wind-aware path search and
B\'ezier trajectory optimization~\cite{habib2026wespr}. Recently, GeoWind2Plan jointly optimizes spatial trajectories and speed profiles using predicted 3D wind and a rotor-power model~\cite{qin2026geowind2plan}.

Existing studies have considered wind in formation configuration selection and continuous single-drone trajectory optimization. Extending wind-aware planning to continuously deformable formations introduces coupling among drone trajectories through shared deformation variables. These variables affect both individual drone velocities and the positions where local wind is sampled, jointly influencing air-relative motion cost. Our method incorporates this coupling into a cost evaluated over the entire continuous trajectory family and jointly optimizes formation-center motion, shared affine deformation, and flight time within a virtual tube.

%% file: sections/formulation.tex
\section{Preliminaries and Problem Formulation}\label{sec:formulation}
\subsection{Formation Representation and Virtual Tube}

We represent formation motion and deformation using a
shared affine transformation. Let $\Xi_N\subset\mathbb R^2$
denote a convex $N$-gon with its area centroid at the origin,
where vertex $\xi_i$ corresponds to drone $i$.
To describe both formation geometry and individual drone
motion, we apply the shared affine transformation to
any point $\xi\in\Xi_N$, defining a continuous
trajectory family $p(t,\xi)$:
\begin{equation}
p(t,\xi)=p_c(t)+E(t)\xi,\quad
E(t)\in\mathbb R^{3\times2},
\label{eq:family}
\end{equation}
where $p_c(t)\in\mathbb R^3$ is the formation-center
trajectory and $E(t)$ is the time-varying affine
deformation matrix governing formation orientation,
scale, and shape. The trajectory of drone $i$ is
obtained by setting $\xi=\xi_i$:
\begin{equation}
p_i(t)=p(t,\xi_i)=p_c(t)+E(t)\xi_i,
\quad i=1,\ldots,N.
\label{eq:drone}
\end{equation}

To constrain the formation to free flight space, we use a shared virtual tube $\mathcal T$ composed of $M$ safe spheres $\mathcal B_m$:
\begin{equation}
\begin{aligned}
\mathcal T&=\bigcup_{m=1}^{M}\mathcal B_m,\\
\mathcal B_m&=\{p\in\mathbb R^3:\|p-c_m\|\le r_m\},
\end{aligned}
\label{eq:tube}
\end{equation}
where $c_m$ and $r_m$ are the sphere center and effective
radius after accounting for safety margins. To enforce tube constraints piecewise and optimize flight-time allocation, the trajectory is divided into $M$ pieces following the sphere sequence. Each piece $m$ is assigned to $\mathcal B_m$ with duration $T_m>0$, and the total flight time is $T_f=\sum_{m=1}^{M}T_m$. The planned virtual tube and drone trajectories in the Chicago scene are shown in Fig.~\ref{fig:workflow}.

\subsection{Local Wind and Formation Deformation}
We assume a steady wind field during trajectory planning,
with wind velocity data provided on a discrete 3D CFD grid.
The native wind velocity $w(p)\in\mathbb R^3$ at position
$p$ is obtained by trilinear interpolation of the grid
velocities. To enable continuous optimization involving
wind velocities and their spatial gradients, a smooth
wind field $w_s$ is reconstructed using local cubic
B-spline quasi-interpolation. The resulting field is
$C^2$ continuous within the valid query domain.
In the subsequent simulations, the smoothed field $w_s$ is used for optimization, and the native field $w$ is used for a posteriori evaluation.

From \eqref{eq:family}, the ground-relative velocity is given by
\begin{equation}
\dot p(t,\xi)=\dot p_c(t)+\dot E(t)\xi.
\label{eq:groundvelocity}
\end{equation}
For trajectory optimization, the air-relative velocity
is evaluated using the smoothed wind field:
\begin{equation}
\begin{aligned}
v_a(t,\xi)={}&\dot p(t,\xi)-w_s\!\left(p_c(t)+E(t)\xi\right).
\end{aligned}
\label{eq:airvelocity}
\end{equation}
Equations~\eqref{eq:groundvelocity} and
\eqref{eq:airvelocity} show that formation deformation
affects ground-relative velocity through $\dot E(t)\xi$
and local wind sampling positions through $E(t)\xi$.

\subsection{Problem Formulation}

Given the virtual tube, we jointly optimize the
formation-center trajectory $p_c$, deformation matrix $E$,
and piece durations $\mathbf T=(T_1,\ldots,T_M)$.
As shown in \eqref{eq:planning_problem}, the objective combines four normalized costs:
trajectory smoothness $\widetilde J_s$,
flight time $T_f/T_\star$,
deformation regularization $\widetilde J_E$,
and air-relative motion $\widetilde J_{\rm air}^{s}$,
with corresponding nonnegative weights
$\lambda_s$, $\lambda_T$, $\lambda_E$, and $\lambda_w$.
The flight-time term uses the total duration $T_f$
and a fixed reference duration $T_\star>0$.

The formation must remain within the virtual tube,
with ground-relative speed and acceleration limited by
$v_{\max}$ and $a_{\max}$, and inter-drone distances
restricted to $[d_{\min},D_{\max}]$.
Air-relative speed is bounded by
$v_{a,\max}$ at sampled times and each piece duration lies within
$[T_{\min},T_{\max}]$.
The formation-center trajectory $p_c(t)$ and deformation
matrix $E(t)$ also satisfy the given endpoint conditions.
Hence, the planning problem is formulated as
\begin{equation}
\begin{aligned}
\min_{p_c,E,\mathbf T}\quad&
 \lambda_s\widetilde J_s
 +\lambda_T\frac{T_f}{T_\star}+\lambda_E\widetilde J_E
 +\lambda_w\widetilde J_{\rm air}^{s}\\
\mathrm{s.t.}\quad&
 p(t,\xi)\in\mathcal B_m,\\
&\|\dot p(t,\xi)\|\le v_{\max},\\
&\|\ddot p(t,\xi)\|\le a_{\max},\\
&d_{\min}\le\|p_i(t)-p_j(t)\|\le D_{\max},\\
&\|v_a(t,\xi_i)\|\le v_{a,\max},\\
&T_{\min}\le T_m\le T_{\max},
\end{aligned}
\label{eq:planning_problem}
\end{equation}
where tube containment and ground-relative kinematic
constraints apply to all $\xi\in\Xi_N$ throughout
each assigned piece, and inter-drone distance constraints
apply to all $1\le i<j\le N$ throughout the flight.
The air-relative speed constraint is evaluated at sampled times
at the drone positions.

%% file: figures/workflow.tex
\begin{figure}[t]
\centering
\includegraphics[width=\columnwidth]{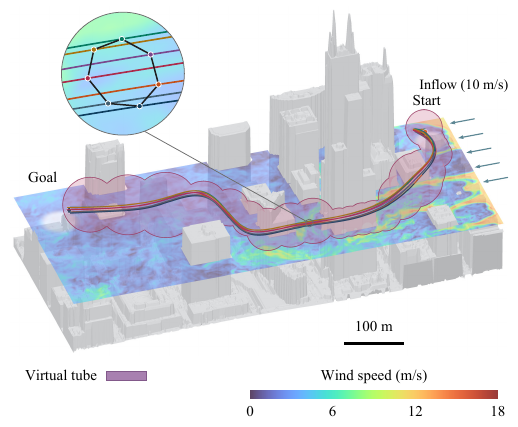}
\caption{Planned seven-drone FA trajectories within
the virtual tube in Chicago. Differently colored
curves represent the trajectories of drones. The enlarged
view shows the local formation configuration.}
\label{fig:workflow}
\end{figure}

%% file: sections/method.tex
\section{Tube-Constrained Trajectory Optimization}\label{sec:method}

The planning problem in Section~\ref{sec:formulation} couples formation
motion and deformation through the wind encountered
at different positions across the formation.
In this section, we first construct a virtual tube that provides safe flight space for the formation. The air-relative motion cost is then formulated, and its wind-dependent variation is derived.
Finally, a compact trajectory parameterization and
a finite set of sufficient constraints convert the
continuous formation planning problem into a
finite-dimensional nonlinear program.

\subsection{Virtual-Tube Construction}

To enable the entire formation to safely traverse cluttered environments, the virtual tube must provide continuous shared flight space with sufficient overlap between adjacent safe spheres. Inspired by Tube RRT*~\cite{mao2025tuberrt}, a safe-sphere roadmap is constructed from obstacle geometry. Two spheres are connected if their
overlap can accommodate the initial
formation with the prescribed
safety margins. Path length, passage width, the number of spheres,
and changes in heading are jointly considered
when selecting the safe-sphere sequence. More details are provided in the Appendix.

\subsection{Wind--Deformation Coupling and Cost Evaluation}
\label{sec:wind_coupling}
\subsubsection{Air-Relative Motion Cost}
The shared affine map of the formation defines
a trajectory for each $\xi\in\Xi_N$, forming
a continuous trajectory family.
Drone $i$ follows the trajectory associated
with vertex $\xi_i$ of the polygon.
We use squared air-relative speed as a proxy for dynamic-pressure-related aerodynamic loading and define the local motion cost as follows:
\begin{equation}
\begin{aligned}
q_{\mathrm{air}}^s(t,\xi)
&=\|v_a(t,\xi)\|^2\\
&=\left\|
\dot p_c(t)+\dot E(t)\xi
-w_s\bigl(p_c(t)+E(t)\xi\bigr)
\right\|^2.
\end{aligned}
\label{eq:local_air_cost}
\end{equation}
This cost depends on both
deformation-induced velocities and
position-dependent local wind.

Since different positions across the formation
may have different velocities and encounter
different local winds, the formation-level cost
must account for this spatial distribution.
Therefore, we use a uniform area average to
evaluate air-relative motion over the entire
continuous formation region.
Under the shared affine map, this average
can be evaluated over the fixed polygon $\Xi_N$.
By integrating the spatially averaged cost
over the flight duration, the normalized formation-level cost can be formulated as
\begin{equation}
\widetilde J_{\mathrm{air}}^s
=\frac{T_\star}{L_\star^2}
\int_0^{T_f}\int_{\Xi_N}
q_{\mathrm{air}}^s(t,\xi)\,d\mu_N\,dt,
\label{eq:normalized_air_cost}
\end{equation}
where $d\mu_N=dA/|\Xi_N|$ is the normalized
area measure, $dA$ is the area element in
the parameter domain, $|\Xi_N|$ is the area
of the polygon, and $L_\star$ and $T_\star$
are fixed normalization scales for length
and time. However, the local wind term $w_s(p_c(t)+E(t)\xi)$
couples the wind field with the formation
trajectories in (\ref{eq:normalized_air_cost}), making the air-relative motion
cost generally difficult to integrate
analytically. Therefore, numerical quadrature is
applied to both the time and spatial integrals.
First, we apply Gauss--Legendre quadrature in time
by mapping each trajectory piece to $[0,1]$.
The temporal quadrature uses nodes $u_j\in(0,1)$
and positive weights $\omega_j>0$ satisfying
$\sum_j\omega_j=1$.
For piece $m$ of duration $T_m$, the physical
evaluation times are $t_{mj}=t_m^0+T_m u_j$,
with $t_m^0=\sum_{h=1}^{m-1}T_h$,
and the corresponding weights are $T_m\omega_j$. Then, for spatial integration, we triangulate
$\Xi_N$ and apply Duffy-transformed Gaussian
quadrature to each triangle~\cite{duffy1982quadrature}.
This resulting spatial nodes $\zeta_k\in\Xi_N$
and positive weights $\nu_k$, where the weights
are normalized such that $\sum_k\nu_k=1$.
Finally, the discrete air-relative motion cost is
\begin{equation}
\widehat{\widetilde J_{\mathrm{air}}^s}
=\frac{T_\star}{L_\star^2}
\sum_{m=1}^{M}T_m
\sum_{j=1}^{n_t}\sum_{k=1}^{n_\xi}
\omega_j\nu_k\,
q_{\mathrm{air}}^s(t_{mj},\zeta_k).
\label{eq:quadrature}
\end{equation}

For the subsequent trajectory optimization, it is necessary to obtain the derivatives of the air-relative motion cost.  At fixed time $t$ and coordinate $\xi$, the local first-order variation is
\begin{equation}
\begin{gathered}
\delta q_{\mathrm{air}}^s
=2(v_a)^\top\delta v_a,\\
\begin{aligned}
\delta v_a
={}&\delta\dot p_c+\delta\dot E\xi-\nabla w_s(p_c+E\xi)
(\delta p_c+\delta E\xi).
\end{aligned}
\end{gathered}
\label{eq:variation}
\end{equation}
The first-order change in local wind velocity
is given by the product of the spatial Jacobian
$\nabla w_s(p_c+E\xi)$ and the position
perturbation $\delta p_c+\delta E\xi$. These local variations are then used to construct the quadrature-based gradient of the air-relative motion cost.

\subsubsection{Trajectory Smoothness Cost}
In this work, trajectory smoothness is measured by snap. Averaging the squared snap over the entire continuous trajectory family gives the smoothness cost:
\begin{equation}
\begin{aligned}
J_s
&=\int_0^{T_f}\int_{\Xi_N}
\|p^{(4)}(t,\xi)\|^2\,d\mu_N\,dt\\
&=\int_0^{T_f}\left(
\|p_c^{(4)}(t)\|^2+
\operatorname{tr}\left(
E^{(4)}(t)\Sigma_NE^{(4)}(t)^\top
\right)\right)dt,
\end{aligned}
\label{eq:smoothness_cost}
\end{equation}
where $\Sigma_N=\int_{\Xi_N}\xi\xi^\top\,d\mu_N$ is the second-moment matrix of the polygon. Since the area centroid of $\Xi_N$ is at the origin, the cross term vanishes upon area averaging. The normalized cost is $\widetilde J_s=(T_\star^7/L_\star^2)J_s$.

\subsubsection{Deformation Regularization Cost}
The deformation regularizer penalizes deviations from a reference configuration. A fixed reference affine map $E_{\mathrm{ref}}\in\mathbb R^{3\times2}$ defines the nominal shape, scale, and orientation of the formation. The normalized deformation cost is
\begin{equation}
\widetilde J_E
=\frac{1}{a_\star^2T_\star}
\int_0^{T_f}
\|E(t)-E_{\mathrm{ref}}\|_F^2\,dt,
\label{eq:deformation_cost}
\end{equation}
where $a_\star=\ell_{\mathrm{nom}}/\rho_N$, $\ell_{\mathrm{nom}}$ is the nominal nearest-neighbor distance, and $\rho_N=\min_{i<j}\|\xi_i-\xi_j\|$ is the minimum distance between polygon vertices.

\subsection{Constraint Enforcement and Trajectory Optimization}

The joint optimization must satisfy geometric and
kinematic constraints throughout each trajectory piece.
Tube containment and ground-relative speed and acceleration bounds are imposed on the Bernstein control points of the vertex trajectories and their derivatives. The maximum inter-drone distance is bounded using pairwise relative-position control points. The Bernstein convex-hull property ensures these bounds throughout each trajectory piece~\cite{park2022lsc}. Finally, the affine dependence on $\xi$ and the convexity of $\Xi_N$ extend tube containment and kinematic bounds from the vertices to the entire continuous trajectory family.

Unlike the upper bounds, minimum inter-drone separation is a nonconvex distance lower bound that cannot be directly guaranteed by the Bernstein convex-hull property. Under the shared affine representation, the relative position of any two drones satisfies $p_i(t)-p_j(t)=E(t)(\xi_i-\xi_j)$. Following the minimum-stretch condition for affine formations~\cite{rastgoftar2022continuum}, the deformation matrix is constrained by
\begin{equation}
E(t)^\top E(t)\succeq s_{\mathrm{req}}^2I_2,
\qquad s_{\mathrm{req}}=\frac{d_{\min}}{\rho_N}.
\label{eq:spectral}
\end{equation}
This condition ensures $\|p_i(t)-p_j(t)\|\ge s_{\mathrm{req}}\|\xi_i-\xi_j\|\ge d_{\min}$, satisfying the minimum separation constraint.

To enforce this condition throughout each trajectory piece, the matrix polynomial of piece $m$ is expressed in Bernstein form
\begin{equation}
\frac{E_m(u)^\top E_m(u)}{s_{\mathrm{req}}^2}-I_2
=\sum_{\ell=0}^{2d}G_{m,\ell}B_\ell^{2d}(u),
\label{eq:matrix_bernstein}
\end{equation}
where $u\in[0,1]$ is the normalized piece time,
$E_m(u)=E(t_m^0+T_mu)$,
$d$ is the polynomial degree, $B_\ell^{2d}(u)$ denotes the Bernstein basis, and $G_{m,\ell}$ are symmetric coefficient matrices. Since the Bernstein basis functions are
nonnegative and form a partition of unity,
requiring all $G_{m,\ell}\succeq 0$ provides
a sufficient condition for continuous-time
minimum separation.

To jointly optimize formation-center motion,
deformation, and timing within the selected virtual
tube, the shared variables are represented by a
nine-dimensional trajectory
$y(t)=[p_c(t)^\top,e_1(t)^\top,e_2(t)^\top]^\top
\in\mathbb R^9$, where $E(t)=[e_1(t),e_2(t)]$.
To reduce the number of optimization variables
and retain the preceding Bernstein-based
continuous-time constraints, we adopt the
seventh-degree minimum-snap MINCO
parameterization~\cite{wang2022minco}.
Given the intermediate states $Q\in\mathbb R^{(M-1)\times9}$, piece durations  $\mathbf T=(T_1,\ldots,T_M)$, and specified boundary states, MINCO reconstructs a $C^6$-continuous piecewise-polynomial trajectory, which is then converted to Bernstein form for constraint evaluation.
Thus, the optimization involves $10M-9$
scalar decision variables in $(Q,\mathbf T)$,
without independently optimizing the
B\'ezier control points.
This compact representation reduces the
decision dimension while restricting the search
to the MINCO trajectory family.

Finally, sequential least squares programming (SLSQP) is used to optimize $(Q,\mathbf T)$, with the objective gradient and constraint Jacobians computed using MINCO coefficient derivatives and the wind-field variations derived in Section~\ref{sec:wind_coupling}. After optimization, the solution is checked
for constraint feasibility and first-order
stationarity. The air-relative motion cost is then reevaluated
using the original CFD wind field, and the
accuracy of numerical integration is verified.

%% file: sections/experiments.tex
\section{Simulations and Results}
\label{sec:experiments}\label{sec:results}

To evaluate the proposed wind-aware planner, we
use two ablation variants to examine how the
wind-aware objective and local wind evaluation
affect air-relative motion cost, flight time,
closed-loop tracking accuracy, and adverse-wind
exposure across four urban scenes.
We further investigate the influence of the
wind-cost weight $\lambda_w$ on these outcomes.
Finally, a comparison with independently
parameterized drone trajectories evaluates the
computational benefit of the shared affine
representation.

\subsection{Simulation Setup}
\subsubsection{Simulation Environments and Ablation Variants}

\input{figures/geometry}

We evaluate our method in four urban scenes, as shown in
Fig.~\ref{fig:geometry}. Cross involves a turn at an intersection,
Street traverses a narrow urban canyon, and Mixed follows a longer
route through buildings of varying heights. Chicago extends the
evaluation to an urban scene reconstructed from LiDAR point cloud
data\footnote{Illinois LiDAR data: \url{https://clearinghouse.isgs.illinois.edu/data/elevation/lidar/}}. The urban wind fields are computed using
FluidX3D~\cite{lehmannfluidx3d}, with an inlet
speed of 10~m/s in all four scenes.

In this work, we design two ablation variants to isolate the effects of the wind objective and
spatial wind evaluation. Wind-objective-off (WOO) removes the
air-relative motion cost from the objective but retains the air-relative speed constraint. Center-wind shared kinematics (CWSK)
retains the air-relative motion cost and deformation-induced drone
velocities but evaluates wind only at the formation center.
By contrast, Family-Air (FA) uses the full objective with local wind evaluated
across the moving formation. All three methods otherwise use the same
shared affine representation, virtual tube, task conditions,
solver settings, and numerical quadrature protocol. Table~\ref{tab:design} lists the benchmark settings
and numerical acceptance thresholds, including
the tube center-chain length $L_{\mathcal T}$
and CFD grid spacing $\Delta x$. Trajectory optimization is performed on a computer
equipped with an Apple M5 processor and 24 GB of
unified memory, using the SLSQP implementation
in SciPy.

\subsubsection{Closed-Loop Simulation Setup}

To evaluate the dynamic execution of the
optimized trajectories, we use a
six-degree-of-freedom quadrotor model
with mass $m=1.5\,\mathrm{kg}$,
inertia matrix $J=\operatorname{diag}
(0.025,0.025,0.045)\,\mathrm{kg\,m^2}$,
and a geometric proportional--derivative
controller. The simulated position of drone $i$ is denoted
by $p_i^{\rm sim}(t)$.
Aerodynamic drag is modeled as
$-\tfrac12\rho C_dA_f\|v_{\rm rel}\|v_{\rm rel}$,
where $v_{\rm rel}=\dot p_i^{\rm sim}
-w(p_i^{\rm sim})$ is the air-relative
velocity, $\rho=1.225$~kg/m$^3$ is the
air density, and $C_dA_f=0.060$~m$^2$
is the product of the drag coefficient
and frontal area. 

\subsubsection{Evaluation Metrics}

For each reference coordinate $\xi\in\Xi_N$,
the unweighted air-relative motion cost is
$\mathcal A(\xi)$. The cost metrics are
\begin{equation}
\begin{aligned}
\mathcal A(\xi)
  &=\int_0^{T_f}
  \|\dot p(t,\xi)-w(p(t,\xi))\|^2\,dt,\\
J_{\rm fam}
  &=\int_{\Xi_N}\mathcal A(\xi)\,d\mu_N,\\
\bar J_N
  &=\frac1N\sum_{i=1}^{N}\mathcal A(\xi_i),\quad
J_{\rm worst}=\max_{1\le i\le N}\mathcal A(\xi_i),
\end{aligned}
\label{eq:results_metrics}
\end{equation}
where $J_{\rm fam}$ is the area-averaged
air-relative motion cost over the reference
polygon, $\bar J_N$ is the mean cost of the
individual drones, and $J_{\rm worst}$ is
the largest individual-drone cost.

Closed-loop tracking accuracy is evaluated
using 
\begin{equation}
\begin{aligned}
e_i(t)
 &=\|p_i^{\rm sim}(t)-p_i(t)\|,\\
\bar e_{\rm RMS}
 &=\frac1N\sum_{i=1}^{N}
 \left(\frac1{T_f}\int_0^{T_f}
 e_i^2(t)\,dt\right)^{1/2},\\
e_{\max}
 &=\max_{i,k}e_i(t_k),
\end{aligned}
\label{eq:tracking_metrics}
\end{equation}
where $e_i(t)$ is the position tracking
error of drone $i$, $\bar e_{\rm RMS}$
averages the RMS errors of individual
drones, and $e_{\max}$ is the largest
sampled position error over all drones,
with $t_k$ denoting the saved simulation times.

We also evaluate the adverse-wind exposure
fraction and its cumulative duration:
\begin{equation}
\begin{aligned}
R_{\rm adv}
 &=\frac{1}{NT_f}\sum_{i=1}^{N}
 \int_0^{T_f}h_i(t)\,dt,\\
h_i(t)
 &=\mathbf{1}\left[
 \|v_{a,i}(t)\|^2>\|\dot p_i(t)\|^2
 \right],\\
\tau_{\rm adv}
 &=T_f R_{\rm adv},
\end{aligned}
\label{eq:wind_exposure_metrics}
\end{equation}
where $v_{a,i}(t)=\dot p_i(t)-w(p_i(t))$,
and $h_i(t)$ indicates whether local wind
increases the air-relative speed compared
with wind-free motion at the same ground speed.
Specifically, $R_{\rm adv}$ measures the fraction of flight
time exposed to this condition, and
$\tau_{\rm adv}$ is the corresponding mean
cumulative duration per drone.

\subsection{Planning Performance}
\label{sec:wind_planning_results}

\subsubsection{Air-Relative Motion Cost and Flight Time}
\label{sec:air_cost_comparison}
Figure~\ref{fig:contrasts}(a) compares the
air-relative motion costs of WOO, CWSK,
and FA, averaged over ten planning tasks
with different starting positions in each scene.
FA reduces $J_{\rm fam}$ by 22.14--48.53\%
relative to WOO and by 2.09--15.76\%
relative to CWSK.
These reductions demonstrate the benefit
of incorporating the air-relative motion
cost and evaluating local wind across
the moving formation.
The mean-drone and worst-drone costs
are also lower with FA on average in all
four scenes.
\input{figures/contrasts}

The lower air-relative motion costs are
accompanied by longer flight times in
Cross, Mixed, and Chicago.
Relative to WOO, flight duration increases
by 9.29\%, 24.88\%, and 27.96\%,
respectively, and remains nearly unchanged
in Street.
Longer durations can reduce air-relative
speeds along parts of the trajectory,
balancing the wind-related cost against
the flight-time penalty.

\subsubsection{Tracking Accuracy and Wind Exposure}
\label{sec:trajectory_tracking}

Using the same planning tasks in Sec.~\ref{sec:air_cost_comparison},
we assess closed-loop tracking errors
and adverse-wind exposure along the
planned trajectories. The results are
shown in Fig.~\ref{fig:contrasts}(b).
It can be seen that FA reduces the mean RMS position error
by 13.88--53.27\% relative to WOO
and by 1.46--15.41\% relative to CWSK.
The scene-averaged peak errors show
the same trend across all four scenes.
At the nominal wind-cost weight, FA reduces both
the exposure fraction $R_{\rm adv}$ and
the mean cumulative exposure time
$\tau_{\rm adv}$ relative to WOO and CWSK
across all four scenes.
Compared with WOO, $R_{\rm adv}$ decreases
by 12.17--45.30\%, and
$\tau_{\rm adv}$ decreases by 2.61--13.39~s.

\subsubsection{Wind-Weight Sensitivity}

To examine how the wind-cost weight affects
planning and tracking performance, we evaluate
$q=\lambda_w/\lambda_w^{\rm nom}\in\{0,0.5,1,2\}$,
where $q=0$ corresponds to WOO with the air-relative speed constraint unchanged. Figure~\ref{fig:sensitivity} shows the
cost reduction $\Delta J=100(1-J_{\rm fam}/J_{\rm fam}^{\rm WOO})$,
the mean RMS tracking error $\bar e_{\rm RMS}$,
and the adverse-wind exposure metrics
$R_{\rm adv}$ and $\tau_{\rm adv}$.

$J_{\rm fam}$ and the mean RMS tracking error decrease as $q$ increases across all four scenes.
Although tracking error is not directly
included in the planning objective,
its reduction under the same controller
indicates improved closed-loop tracking
performance. These improvements are
accompanied by longer flight durations
in Cross, Mixed, and Chicago, with only
minor changes in Street. The adverse-wind exposure metrics reveal
a different trend. As $q$
increases, $R_{\rm adv}$ decreases in
Cross, Street, and Mixed, but rises slightly
in Chicago from $q=1$ to $q=2$. Because $\tau_{\rm adv}=T_fR_{\rm adv}$,
a lower exposure fraction does not
necessarily imply shorter cumulative
exposure time.
\input{figures/sensitivity}

\subsection{Computational Efficiency}
\label{sec:nsetup}
\label{sec:nresults}

The WOO, CWSK, and FA methods use a common
affine trajectory representation, referred
to as Shared in the following comparison.
To assess its computational benefit, we
introduce a Joint baseline that assigns
an independent seventh-degree MINCO
trajectory to each drone under a matched
wind-aware objective. Both representations evaluate air-relative
motion cost over the reference polygon.
The Joint baseline interpolates the drone trajectories
within triangular regions of this polygon,
allowing more flexible deformation than
the single affine map used by Shared.
For a virtual tube with $M$ pieces, Joint
optimizes $M-1$ three-dimensional intermediate
waypoints for each drone and $M$ shared
segment durations.
This results in $3N(M-1)+M$ decision variables,
compared with $10M-9$ for Shared.
Consequently, the two representations have the same number
of variables at $N=3$, but Shared uses fewer
variables as $N$ increases.
\input{tables/shared_joint}
Table~\ref{tab:shared_joint} reports the
Joint-to-Shared ratios of active SLSQP
and method-process times. The former excludes
preparation and verification pauses, and the
latter includes data loading, derivative
preparation, optimization, and numerical
acceptance checks.
Let $F_{\rm Shared}$ and $F_{\rm Joint}$
be the optimized full-objective values
of the two representations.
Their relative difference is
$\Delta F=100(F_{\rm Shared}/F_{\rm Joint}-1)$,
with positive values favoring Joint. Shared achieves shorter SLSQP times in
19 of the 20 comparisons and shorter
method-process times in all 20,
with median speedups of $3.35\times$
and $2.45\times$. The median relative objective difference
is only $+0.05\%$. These results demonstrate the computational
advantage of the shared affine representation
on the evaluated tasks, without a systematic
difference in the achieved objective values.

\FloatBarrier

%% file: figures/geometry.tex
\begin{figure}[!t]
\centering
\marginSafeFigure[0.88\columnwidth]{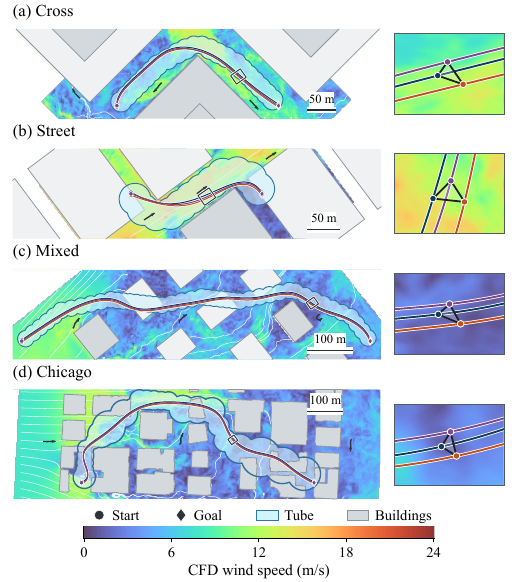}
\caption{Benchmark setups: representative urban CFD wind fields
and three-drone FA trajectories in (a) Cross, (b) Street,
(c) Mixed, and (d) Chicago. Differently colored
curves represent the trajectories of drones. The right panels show enlarged views
of the highlighted regions.}
\label{fig:geometry}
\vspace{5pt}
\begingroup
\makeatletter
\def\@captype{table}
\makeatother
\input{tables/design}

\endgroup
\end{figure}

%% file: tables/design.tex
\caption{Benchmark settings and numerical acceptance thresholds}
\label{tab:design}
\centering
\scriptsize
\renewcommand{\arraystretch}{1.12}

\begin{tabular*}{\columnwidth}{
@{}l@{\hspace{5pt}}c@{\hspace{7pt}}c
@{\extracolsep{\fill}}c@{\hspace{6pt}}c@{}}
\toprule
Scene & $M$
& $L_{\mathcal T}=\sum_{m=1}^{M-1}
\|c_{m+1}-c_m\|$ (m)
& CFD grid & $\Delta x$ (m) \\
\midrule
Cross   & 22 & 335.05 & $815\times815\times401$ & 0.798 \\
Street  & 12 & 229.44 & $685\times1095\times356$ & 0.730 \\
Mixed   & 44 & 819.35 & $805\times966\times343$ & 0.933 \\
Chicago & 24 & 806.61 & $732\times732\times499$ & 1.503 \\
\midrule

\multicolumn{5}{@{}l@{}}{
\textit{Common constraints and nominal settings}}\\
\addlinespace[2pt]
\multicolumn{5}{@{}l@{}}{%
\begin{tabular*}{\columnwidth}{
@{}l@{\extracolsep{\fill}}l@{}}
$d_{\min}=3.4$ m,\ $D_{\max}=6.5$ m ($N=3$)
& $v_{\max}=10$ m/s,\ $a_{\max}=8$ m/s$^2$\\
$v_{a,\max}=22$ m/s
& $1\le T_m\le15$ s\\
$L_\star=10$ m,\ $T_\star=4$ s
& $a_\star=4.8$ m\\
$\lambda_s=0.02,\ \lambda_T=1$
& $\lambda_E=0.10,\ \lambda_w=0.15$\\
\end{tabular*}
}\\
\midrule

\multicolumn{5}{@{}l@{}}{
\textit{Numerical acceptance thresholds}}\\
\addlinespace[2pt]
\multicolumn{5}{@{}l@{}}{%
\begin{tabular*}{\columnwidth}{
@{}l@{\extracolsep{\fill}}l@{}}
Stationarity $\le 10^{-3}$
& Primal feasibility $\le 10^{-6}$\\
Complementarity $\le 2\times10^{-5}$
& Dual feasibility $\le 10^{-12}$\\
\end{tabular*}
}\\

\bottomrule
\end{tabular*}

%% file: figures/contrasts.tex
\begin{figure}[t]
\centering
\includegraphics[width=\columnwidth]{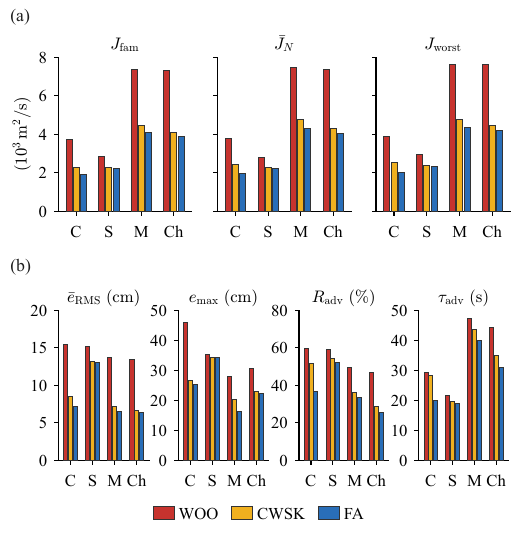}
\caption{
Comparison of WOO, CWSK, and FA
across Cross (C), Street (S), Mixed (M),
and Chicago (Ch).
(a) Air-relative motion costs:
$J_{\rm fam}$, $\bar J_N$, and $J_{\rm worst}$.
(b) Tracking errors $\bar e_{\rm RMS}$
and $e_{\max}$, and adverse-wind exposure
metrics $R_{\rm adv}$ and $\tau_{\rm adv}$. All results are averaged over ten planning tasks per scene.}
\label{fig:contrasts}
\end{figure}

%% file: figures/sensitivity.tex
\begin{figure}[t]
\centering
\marginSafeFigure[\columnwidth]{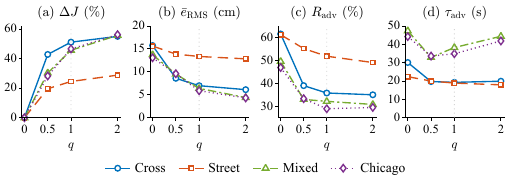}
\caption{Wind-weight sensitivity across four
urban scenes.
(a) Air-relative motion cost reduction
$\Delta J$ relative to WOO.
(b) Mean RMS tracking error $\bar e_{\rm RMS}$.
(c) Adverse-wind exposure fraction $R_{\rm adv}$.
(d) Mean cumulative exposure time $\tau_{\rm adv}$.
The zero-weight case ($q=0$) corresponds to
WOO, and the remaining weights use FA.}
\label{fig:sensitivity}
\end{figure}

%% file: tables/shared_joint.tex
\begin{table}[t]
\caption{Shared--Joint comparison}
\label{tab:shared_joint}
\centering
\begingroup
\scriptsize
\setlength{\tabcolsep}{2.2pt}
\renewcommand{\arraystretch}{1.03}
\begin{tabular}{@{}llrrrrr@{}}
\toprule
Scene & Metric & $N=3$ & 4 & 5 & 6 & 7 \\
\midrule
Cross
 & SLSQP & 2.09 & 1.16 & 3.13 & 4.90 & 14.09 \\
 & Process & 4.04 & 2.50 & 3.52 & 4.54 & 4.98 \\
 & $\Delta F$ (\%) & +0.000 & +0.015 & +0.380 & +0.855 & +0.953 \\
\midrule
Street
 & SLSQP & 1.66 & 0.76 & 5.73 & 2.00 & 4.79 \\
 & Process & 1.97 & 1.08 & 2.40 & 1.19 & 1.18 \\
 & $\Delta F$ (\%) & +0.000 & +0.148 & +0.200 & +0.246 & +0.263 \\
\midrule
Mixed
 & SLSQP & 5.30 & 1.91 & 2.73 & 5.75 & 11.92 \\
 & Process & 10.18 & 1.63 & 1.18 & 2.55 & 3.36 \\
 & $\Delta F$ (\%) & -4.061 & +0.058 & +0.043 & +0.085 & +0.013 \\
\midrule
Chicago
 & SLSQP & 1.20 & 1.22 & 3.57 & 4.84 & 9.28 \\
 & Process & 1.98 & 1.40 & 1.35 & 2.56 & 3.17 \\
 & $\Delta F$ (\%) & +1.005 & -2.115 & +0.023 & -1.057 & -2.873 \\
\bottomrule
\end{tabular}
\endgroup
\end{table}

%% file: sections/conclusion.tex
\section{Conclusion}
\label{sec:conclusion}

This work presents Family-Air (FA), a wind-aware trajectory planner for deformable aerial formations within virtual tubes. The proposed wind–deformation coupling formulation evaluates air-relative motion cost over a continuous trajectory family, accounting for both deformation-induced drone velocities and position-dependent local wind. FA jointly optimizes formation-center motion, shared affine deformation, and segment durations using a compact MINCO parameterization with Bernstein-based sufficient conditions for continuous-time constraints. Simulations across four urban scenes demonstrate reduced air-relative motion costs and adverse-wind exposure, together with improved closed-loop tracking performance compared with the WOO and CWSK variants. The shared affine representation also improves computational efficiency compared with independently parameterized drone trajectories. Future work will incorporate propulsion-power models to evaluate actual energy consumption.